\documentclass[journal]{IEEEtran}
\usepackage{graphicx}
\usepackage{amsmath}
\usepackage{amsfonts}
\usepackage{tabularx}
\usepackage{multirow}
\usepackage[caption=false,font=normalsize,labelfont=sf,textfont=sf]{subfig}
\usepackage{amsthm}
\usepackage{url}
\usepackage{enumitem}

\DeclareMathOperator*{\argmin}{arg\,min}
\DeclareMathOperator*{\argmax}{arg\,max}

\newcommand{\bs}[1]{\boldsymbol #1}

\usepackage{color}

\newcommand{\bx}{{\bf x}}
\newcommand{\by}{{\bf y}}

\newcommand{\bX}{{\bf X}}
\newcommand{\bY}{{\bf Y}}

\begin{document}

\title{External Patch-Based Image Restoration Using Importance Sampling}

\author{Milad Niknejad, Jos\'e  Bioucas-Dias, M\'ario A.T. Figueiredo
\thanks{The authors are with the Instituto de Telecomunica\c{c}\~oes and Instituto Superior T\'ecnico, University of Lisbon, Portugal.}
\thanks{The research leading to these results has received funding from the European Union's 
Seventh Framework Programme (FP7-PEOPLE-2013-ITN) under grant agreement n° 607290 SpaRTaN.}}

\markboth{}%
{Niknejad \MakeLowercase{\textit{et al.}}: External Patch-Based Image Restoration Using Importance Sampling}

\maketitle

\begin{abstract}
This paper  introduces a new approach to patch-based  image restoration based on  external datasets and  importance sampling. The \textit{minimum mean squared error} (MMSE)  estimate of the  image patches, the computation of which requires solving a multidimensional (typically intractable) integral, is approximated using samples from an external dataset. The new method, which can be interpreted as a generalization of the external non-local means (NLM), uses self-normalized importance sampling to efficiently approximate the MMSE estimates. The use of self-normalized importance sampling endows the proposed method with great flexibility, namely regarding the statistical properties of the measurement noise.  The effectiveness of the  proposed method is shown in a series of experiments using both generic large-scale and class-specific external datasets.
\end{abstract}

\IEEEpeerreviewmaketitle

\section{Introduction}
Under the framework of imaging inverse problems, image restoration aims at reverting the degradation introduced by the image acquisition process. The observation is often modeled as a linear function of the  underlying original image, contaminated by noise. Formally, with $\mathbf{x} \in \mathbb{R}^n$ being the vector whose entries are the (lexicographically ordered) pixel values of the original image, and $\mathbf{y} \in \mathbb{R}^m$ denoting the observed image, the linear model is written as
\begin{equation}
\label{eq:LIP}
\mathbf{y}=\mathbf{H} \mathbf{x}+\mathbf{v},
\end{equation}
where $\mathbf{v}$ models additive noise (often zero-mean, independent,  Gaussian  distributed), and ${\bf H}\in \mathbb{R}^{m\times n}$ is a matrix modeling  the  observation/degradation process (blur, convolution, projection, etc...). To address the image restoration problem using Bayesian tools, a central building block is the translation of the forward model \eqref{eq:LIP} into a conditional probability (density or mass) function of $\bf y$, given $\bf x$, also called the \textit{likelihood function}. In the case of Gaussian noise, this yields a well-know conditional probability density function (pdf)
\begin{equation}
f_{Y|X}(\mathbf{y}|\mathbf{x}) = {\cal N}({\bf y}|{\bf Hx},\sigma^2 {\bf I}) \propto \exp\Bigl( -\frac{1}{2\, \sigma^2} \|{\bf y - H\, x} \|_2^2\Bigr).
\label{eq:glik}
\end{equation}
where $\sigma^2$ is the noise variance and $\bf I$ denotes an identity matrix.

In many important cases, the observation model departs from  \eqref{eq:LIP}, namely because there are non-linear effects or the noise is not additive. One such case is when  the observation noise is Poissonian, which is a very large and important class of problems \cite{2014_salmon_poisson,2011_makitalo_optimal,niknejad_poisson_2018}. 
In this case, each observed pixels $({\bf y})_l \in \mathbb{N}_0$, for $l\in 1,\dots,m$, is a realization of a Poisson random variables with mean $(\mathbf{H}\mathbf{x})_{l}$. Under the conditional {independence} assumption, the conditional probability mass function of $\bf y$ given $\bf x$ is thus 
\begin{equation}
\label{eq:poisson}
f_{Y|X}(\mathbf{y}|\mathbf{x}) = \prod_{l=1}^m \frac{e^{-(\mathbf{H}\mathbf{x})_{l}}(\mathbf{H}\mathbf{x})_{l}^{(\mathbf{y})_{l}}}{(\mathbf{y})_{l}!}.
\end{equation}

Image restoration methods are often specified for one type of observation and noise model. A significant amount of work has been devoted to Gaussian denoising, which corresponds to \eqref{eq:LIP} with $\bf H = I$.  Some methods rely only on the noisy image itself \cite{2007_dabov_bm3d,2006_elad_denoisesparse,2005_buades_nonlocalmenas,2013_dong_lssc,Niknejad_TIP2015, Teodoro2015}, on an external dataset of clean images \cite{2014_luo_image,2011_zoran_learning}, or a combination of both \cite{2015_chen_external,2013_mosseri_combining}.  Methods based on  external datasets have been recently an active research topic, especially with the emergence of \textit{deep neural networks} (DNN), which require a large set of training data (see \cite{zhang_beyond_2017,chen_trainable_2017,jiao_formresnet_2017,vemulapalli_deep_2016,remez_deep_2017} and references therein). Most of these methods require separate training for different image restoration tasks and even for different parameters of the observation model, namely the noise variance.  A  different approach, also using external dataset of images, learns some parametric distribution, often \textit{Gaussian mixture models} (GMMs), to the image patches  \cite{2011_zoran_learning,2015_chen_external}. The learned models are then used to  regularize the restored images. However, these methods have been mostly limited to linear inverse problems with Gaussian noise, due to the difficulty in obtaining MAP or MMSE estimates for other noise models.

Another major group of image restoration methods based on external datasets works by using sample patches (instead of learning parametric models) to restore the observed patches \cite{2013_mosseri_combining,freeman_example_2002,2011_zontak_internal,adams_gaussian_2009,hays_scene_2007,2014_chan_monte1}. To estimate the central pixel of the patch, most of these methods compute weighted average of central pixels of clean patches (selected from the external dataset). The weights are obtained based on the exponential of negative of a distance between noisy and clean patches. A method from this family for Gaussian denoising is the so-called \textit{external non-local-means} (NLM),  in which the distance is computed by the $\ell_2$ norm of the difference between patches divided by the noise variance  \cite{2011_levin_natural,2012_levin_patch}. This estimate is shown to converge to the MMSE estimate as the number of samples approach infinity \cite{2011_levin_natural}. The name ``external NLM'' is inspired by well-known (internal) NLM \cite{2005_buades_nonlocalmenas}, in which the same weighted averaging is computed, but with the patches from the noisy image itself and the distance in the exponential is divided by a hand-tuned parameter. However, external NLM is computationally demanding, as reported in \cite{2011_levin_natural}, due to the need to use a large external dataset of clean patches \footnote{There are some variants of \cite{2011_levin_natural}  which are also called external NLM. However, in this paper, by external NLM, we mean the method in \cite{2011_levin_natural}, which is shown to converge to the MMSE estimate.}.

In many sampling methods, a selected group of similar patches from the external datasets are used for restoring a patch \cite{freeman_example_2002,2011_zontak_internal,pyatykh_mmse_2015}. It requires computing the distance between the patches of the image to be restored and those of the  external dataset, which, similar to external NLM, in the case of   large-scale  external datasets, involves a huge computational cost. To address this hurdle, some approaches have  focused on  accelerating the patch matching \cite{he_computing_2012,olonetsky_treecann_2012,barnes_generalized_2010,adams_gaussian_2009}. These methods are often heuristic and rely on hierarchical approaches, such as kd-trees, to approximately find the nearest patches based on the $\ell_2$ distance. In \cite{2014_chan_monte1}, a sampling approach called Monte-Carlo NLM (MC-NLM) is proposed, which uses approaches form large deviation theory to speed up the NLM algorithm. However, all these methods deal with Gaussian noise and the generalization to Poisson noise and other inverse problems {have} not been truly addressed.

In the context of image denoising based on external datasets, a recent class of methods has focused on using class-specific  datasets \cite{remez_deep_2017,2015_luo_adaptive,anwar_category_2017,2016_teodoro_image}. These methods exploit the fact that in many cases, the noisy image is known (or can be easily identified) to belong to a certain class, such as face, text, or fingerprint. 

\textit{Importance sampling} (IS) belongs to the Monte-Carlo (MC) family of methods to approximate multi-dimensional expectations/integrals \cite{robert_monte_2004,evans_methods_1995}. It is often used when sampling from a distribution is difficult, or when reducing the variance of MC estimation is required. The general approach is to sample from a different distribution, called the \textit{proposal distribution}, instead of the \textit{target distribution}, and correcting each function value with a weight that depends on the ratio between the two distributions. If the normalization constant  of {the} proposal or {the} target distribution (or both) {is} unknown, a version of IS called \textit{self-normalized IS} (SNIS) can be used. 

In this paper, we propose a new image restoration method based on SNIS, which is applicable to large-scale external datasets and general observation models, including  non-Gaussian noise. The proposed  approach approximates the MMSE estimate of the image patches using SNIS applied to a set of samples from the external dataset. Instead of sampling from the posterior distribution, which is unknown, we use, as the proposal distribution, a {mixture} of densities derived from clustered patches of the external dataset. The mixture distribution is chosen such that it maximizes the similarity to the optimal sampling distribution of SNIS. The method is non-parametric and the samples are directly derived from the dataset. It is applied to both class-specific and generic image datasets. The obtained results achieve state-of-the-art performance under  Poissonian and Gaussian noise, if a dataset of images from the same class is available. The performance of the proposed method is also shown in inpainting problems with different types of noise. In the case of restoration of generic images, the proposed method can be used to accelerate external NLM, thus enabling the use of the proposed method efficiently on large-scale datasets.

The paper is organized as follows.  Section 2 overviews  MC sampling methods for approximating multivariate expectations (integrals), for later use in the  MMSE estimation of image patches. Section  \ref{sec:proposed} describes the  proposed method. Section 4 reports  experimental results on  class-specific image restoration and restoration using large scale datasets. Finally, Section 5 ends the paper with a few concluding remarks and notes on future work.

\section{Importance Sampling}

Let $f: \mathbb{R}^n \, \to \, \mathbb{R}$ be a real-valued function,  $\mathbf{Z}$ a random vector with pdf $p_Z$, often termed \textit{target distribution} in the context of IS. Then, 
\begin{equation}
	\label{eq:exptf}
	\mu=\mathbb{E}[f(\mathbf{Z})]=\int_{\mathbf{z} \in \mathbb{R}^n} f(\mathbf{z}) \; p_Z(\mathbf{z}) \; d\mathbf{z},
\end{equation}
is the expectation of the random variable $ f({\bf Z})$. It is very often the case that, owing to the large dimensionality of $\bf Z$, the computation of \eqref{eq:exptf} using numerical integration techniques is infeasible. A possible solution is to approximate the integral in 	\eqref{eq:exptf} using MC methods. Let $\mathbf{Z}_1,\ldots,\mathbf{Z}_N\sim p_Z$ be a \textit{random sample} of size $N$ following $p_Z$ (i.e., a sequence of $N$  i.i.d. random variables with pdf $p_Z$), and $\widehat{\mu}_N^{MC}$ be the sampling average random variable  
\begin{equation}
\label{eq:plmc}
\widehat{\mu}_N^{MC}=\widehat{\mathbb{E}}^{MC}_N[f(\mathbf{Z})]= \frac{1}{N} \displaystyle\sum_{j=1}^N{f(\mathbf{Z}_{j})}.
\end{equation}
The strong law of large numbers asserts that $\lim_{N\rightarrow\infty} \widehat{\mathbb{E}}^{MC}_N[f(\mathbf{Z})] = \mathbb{E}[f(\mathbf{Z})]$ (almost surely), provided that $\mathbb{E}|f({\bf Z})|< \infty$ \cite[Theorem 5.18]{wasserman_all_2013}.

In the cases that sampling from the distribution $p_Z$ is intractable, or the estimator \eqref{eq:plmc} has a {large} variance for a specific number of samples $N$, the  IS approach stands as a potential alternative. IS generates samples from a tractable distribution $q_Z$, termed \textit{proposal distribution}, and approximates the expectation \eqref{eq:exptf}  by the weighted average
\begin{equation}
\label{eq:is}
\widehat{\mu}_N^{IS}=\widehat{\mathbb{E}}^{IS}_N[f(\mathbf{Z})]=\frac{1}{N} \displaystyle\sum_{j=1}^N{\frac{p_z(\mathbf{Z}_{j})}{q_z(\mathbf{Z}_{j})}f(\mathbf{Z}_{j})},
\end{equation}
where here $\mathbf{Z}_1,\dots,{\bf Z}_N\sim q_Z$, and we are assuming that  $q_Z(\mathbf{z})>0$ whenever $f({\bf z})\, p_Z(\bf {z}) \ne 0$. Similarly to the plain MC estimator, the  estimator \eqref{eq:is} is also consistent and unbiased under the above-mentioned condition \cite{robert_monte_2004,hesterberg_advances_1988}.

When only unnormalized distributions are available, a self-normalized version of IS (SNIS) may be used. Let $\tilde p_Z=b\,  p_Z$ and $\tilde q_Z=c \, q_Z$ be   unnormalized distributions,
where $b>0$ and $c>0$ are unknown constants.  Assuming that $q_Z(\mathbf{z})>0$ whenever $p_Z(\mathbf{z})>0$, then the  expectation in (\ref{eq:exptf}) may be approximated by 
\begin{equation}
\label{eq:snis}
\widehat{\mu}_N^{SNIS}=\widehat{\mathbb{E}}^{SNIS}_N[f(\mathbf{Z})]=\frac{\displaystyle \sum_{j=1}^{N}{f(\mathbf{Z}_j) w(\mathbf{Z}_j)}}{\displaystyle \sum_{j=1}^{N}{w(\mathbf{Z}_j)}}, 
\end{equation}
where $ \mathbf{Z}_1, \ldots, \mathbf{Z}_N \sim q_Z$ and  $w(\mathbf{z})=\frac{\tilde p_Z(\mathbf{z})}{ \tilde q_Z(\mathbf{z})}$ \cite{robert_monte_2004,hesterberg_advances_1988,mcbook} (termed  \textit{importance weights}). Notice that if, instead of 
$w(\mathbf{z})=\frac{\tilde p_Z(\mathbf{z})}{ \tilde q_Z(\mathbf{z})}$, we used $\tilde w(\mathbf{z})=\frac{c\, \tilde p_Z(\mathbf{z})}{b\, \tilde q_Z(\mathbf{z})} = \frac{ p_Z(\mathbf{z})}{ q_Z(\mathbf{z})}$, nothing would change since the factor $c/b$ appears both in the numerator and the denominator of  \eqref{eq:snis}, thus being irrelevant. The SNIS estimator is biased but consistent under the above-mentioned conditions \cite{hesterberg_advances_1988,mcbook};  it converges therefore to the true value of expectation as the number of samples goes to infinity.

The performance of IS depends critically on the proposal distribution $q_Z$. One may, for example, seek for the  proposal distribution  $q_Z^{*}$ that, for a given sample size,  minimizes the mean square error (MSE); that is 
\begin{equation}
\label{eq:opt_q}
q_Z^{*}=\argmin_q \mathbb{E}\bigl[ \|\widehat{\mu}_N(q)-\mu\|_2^2 \bigr],
\end{equation} 
where $\widehat{\mu}_N(q)$  refers to the estimators $\widehat{\mu}_N^{IS}$ or $\widehat{\mu}_N^{SNIS}$,
using the proposal distribution $q$. For the IS estimator \eqref{eq:is}, the solution of  \eqref{eq:opt_q} is (see \cite{kahn_methods_1953})
\begin{equation}
\label{eq:optis}
q_Z^{*}(\mathbf{z}) \propto |f(\mathbf{z})| \; p_Z(\mathbf{z}),
\end{equation}
whereas for SNIS, the optimal sampling density is
\begin{equation}
\label{eq:optsnis}
q_Z^{*}(\mathbf{z}) \propto |f(\mathbf{z})-\mu| \; p_Z(\mathbf{z}).
\end{equation}
These results are obtained via variational calculus  \cite[Ch. 2]{hesterberg_advances_1988} and used in \cite{hesterberg_weighted_1995,mcbook}. 
The distributions $q_Z^{*}$  in \eqref{eq:optis} and \eqref{eq:optsnis} are also the minimizers of the asymptotic variance of the corresponding  IS and SNIS estimators \cite{hesterberg_weighted_1995}. In the case of SNIS, obtaining the optimal sampling distribution \eqref{eq:optsnis} requires knowing $\mu$, the estimation of which is precisely the goal of SNIS, thus the result cannot be directly applied.  

Estimation of the conditional expectations has been proposed using SNIS. Two well-known methods for this purpose are the \textit{population Monte Carlo} (PMC) \cite{cappe_population_2004} and \textit{adaptive IS} (AIS) \cite{cornuet_adaptive_2012}. The proposal distribution in those methods is  a mixture of densities  designed to guarantee  consistency of the SNIS estimator. The parameters of the mixtures are iteratively updated along the iterations, based on the importance weights. However, those  methods require sampling from a specific parametric distribution. In many applications,  training data is available and can be used as samples. In those cases, the exact distribution of the data is unknown, and fitting any parametric distribution would be an approximation to the true distribution. Consequently, it would be more natural to sample directly from the data, rather than from an approximate fitted distribution. In the next section, we propose an SNIS approach that exploits this idea.

%
%

\section{Proposed Method}
\label{sec:proposed}
\subsection{Introduction}
As already mentioned, image restoration based on sampling from a large-scale dataset is computationally complex. In  this section, we propose a method for image restoration based on SNIS, which can be efficiently implemented for large-scale and/or class-specific external datasets.

Let  $\mathbf{Y} \in \mathbb{R}^n$ denote a random vector associated with the noisy image patches, of size $\sqrt{n}\times\sqrt{n}$, $\mathbf{X}\in \mathbb{R}^n$ the random vector associated with the clean patches corresponding to $\bf Y$,  and $\mathbf{y}\in \mathbb{R}^n$ an observed noise patch (\textit{i.e.}, a sample of $\bf Y$). Consider the goal of estimating the central pixel of the clean patch, denoted ${\bf x}_c$ (leaving aside for now the issue of how the patches are extracted and its estimates combined). As estimation criterion, we adopt  the \textit{minimum mean square error} (MMSE), which is well-known to yield the posterior expectation,
\begin{align}
\label{eq:expectation}
\widehat{\bf x}_{c}=\mathbb{E}[{\bf X}_c|\mathbf{Y}=\mathbf{y}]&=\int_{\mathbf{x} \in \mathbb{R}^n} {\bf x}_c \, p_{X|Y}(\mathbf{x}|\mathbf{y}) \,d\mathbf{x},
\end{align}
where $p_{X|Y}(\cdot\,|\mathbf{y})$ denotes the posterior pdf of $\bf X$ given ${\bf Y} = {\bf y}$. The expression \eqref{eq:expectation} has the structure of \eqref{eq:exptf}, where the function $f({\bf x}) = {\bf x}_c$, \textit{i.e.}, $f:\mathbb{R}^n \rightarrow \mathbb{R}$  extracts the central pixel of the patch.

Computing the  integral \eqref{eq:expectation} is, in general,  intractable. In the special case of multivariate Gaussian distribution for $\mathbf{X}$ and Gaussian additive noise, \eqref{eq:expectation} can be computed in closed form. It is, however, often intractable  for non-Gaussian noise and/or other priors than multivariate Gaussian. Assuming that  the posterior $p_{X|Y}$ is known, the above integral can be approximated by using plain MC. However, very often, the posterior distribution $p_{X|Y}$ is  unknown and thus plain MC sampling cannot be applied.  Although sampling from the unknown posterior is not feasible, samples from the  distribution of natural image patches can be efficiently obtained in our case, as they are available in the external dataset. This motives us to use SNIS to approximate \eqref{eq:expectation}.

\subsection{MMSE Estimation by SNIS: Na\"{i}ve Approach}
In order to approximate the MMSE estimate  in \eqref{eq:expectation} with a SNIS estimator, we start by noting that the target distribution is simply the posterior pdf, which, according to Bayes law, is given by
\begin{align}
p_{X|Y}(\mathbf{x}|\mathbf{y}) & = 
 p_{Y|X}(\mathbf{y}|\mathbf{x})p_X(\bx)/ p_Y(\by),
\end{align}
where  $p_{Y|X}$ is the conditional pdf of ${\bf Y}$ given $\bX$, \textit{i.e.}, the likelihood function, $p_{X}$ is the patch prior, and  $p_{Y}$ is marginal pdf of $\bY$. In the sequel, since $\by$ is given, we often use the compact  notation for the likelihood function
\begin{equation}
 l_{\bf y}(\bx) \equiv p_{Y|X}(\mathbf{y}|\mathbf{x}).
\end{equation}

Given $N$ clean patches ${\bf x}_1,...,{\bf x}_N$ assumed to be i.i.d. samples from the patch prior $p_X$, a na\"{i}ve approach to using SNIS to approximate \eqref{eq:expectation} consists in using $p_X$ as the proposal distribution, leading to
\begin{equation}
\label{eq:isavg}
\widehat{\widehat{\bf x}}_{c} = \frac{\displaystyle \sum_{j=1}^{N}{(\mathbf{x}_j)_c \; w(\mathbf{x}_j)}}{\displaystyle \sum_{j=1}^{N}{w(\mathbf{x}_j)}} = \frac{\displaystyle \sum_{j=1}^{N}{(\mathbf{x}_j)_c \; l_{\bf y}(\mathbf{x}_j)}}{\displaystyle \sum_{j=1}^{N} l_{\bf y}(\mathbf{x}_j)} ,
\end{equation}
because the weights are given by 
\begin{eqnarray}
w(\mathbf{x}_j) = \frac{p_{X|Y}(\mathbf{x}_j|\mathbf{y})}{p_X(\mathbf{x}_j)} \propto \frac{p_{Y|X}(\mathbf{y}|\mathbf{x}_j) \, p_X(\mathbf{x}_j)}{p_X(\mathbf{x}_j)} \nonumber \\= p_{Y|X}(\mathbf{y}|\mathbf{x}_j) =  l_{\bf y}(\bx_j ),
\end{eqnarray}
where we assume that the prior $p_X (\mathbf{x}_j)$ is non-zero which is the case since $\mathbf{x}_j$ is a sample derived from $p_X$.

The drawback of this na\"{i}ve SNIS method is that needs a extremely  large number of external patches to yield a decent estimate, since most sampled patches will be very different from the underlying true one, thus the majority of weights will be extremely small. In fact, a central issue in any IS method (including SNIS) is finding a proposal distribution that is not too different from the target one, such that the weights are not very small. In the next subsection, we propose an approach to tackle this issue for the patch-based image denoising problem.

Finally, notice that if the noise is Gaussian, $l_{\bf y}(\bx_j ) = \exp\bigl( - \| {\bf x}_j - {\bf y} \|_2^2/(2\, \sigma^2)\bigr)$, which shows that the patch estimate in standard external NLM methods \cite{2011_levin_natural,2012_levin_patch}  is nothing more than a SNIS approximation of the MMSE estimate; as far as we know, this had not been noticed before.  Moreover, this is obviously suboptimal, since the proposal distribution should be adapted to the target distribution $p_{X|Y}(\cdot\,|\mathbf{\bf y})$, which is only possible if it depends on the observed $\bf y$.

\subsection{MMSE Estimation by SNIS: Proposed Approach}
As explained in the previous subsection, the target distribution is the posterior pdf $p_{X|Y}$, with an unnormalized version thereof being simply  $l_{\bf y} \, p_X$. In this subsection, we propose a proposal distribution that depends on ${\bf y}$, which is needed to allow adapting it to the particular posterior at hand for each ${\bf y}$.

We begin by clustering the external dataset of patches ${\bf x}_1,...,{\bf x}_L$ into $K$ disjoint clusters: $\mathcal{X}_1,...,\mathcal{X}_K$. Let this clustering induce a partition, $R_1,...,R_K$ of $\mathbb{R}^n$ that satisfies $\mathcal{X}_k \subset R_k$, for $k=1,...,K$. Notice that this partition is obviously not unique, but this will be irrelevant for the proposed method. Using this partition, it is possible to re-write the prior $p_X$, of which ${\bf x}_1,...,{\bf x}_L$ are assumed to be i.i.d. samples, under the form of a mixture,
\begin{equation}
p_X({\bf x}) = \sum_{k=1}^K m_k \; g_k({\bf x}),
\end{equation}
where $m_k g_k$ is the restriction of $p_X$ to $R_k$ and 
\begin{equation}
m_k = \int_{R_k} p_X({\bf x})\, d{\bf x}\label{eq:m_k}
\end{equation} 
is the corresponding normalization constant, that is, 
\begin{equation}
g_k({\bf x}) = \frac{1}{m_k} \left\{ \begin{array}{ll} p_X({\bf x}) & \mbox{if} \; {\bf x}\in R_k\\ 
0 & \mbox{if} \; {\bf x}\not\in R_k.
\end{array} \right.
\end{equation}
Naturally, the elements of cluster $\mathcal{X}_k$ are assumed to be samples of a random variable (say $\mathbf{X}_k$) following the pdf $g_k$.

We suggest the proposal distribution herein to be a mixture with components $g_1,...,g_K$ and weights $\alpha_1({\bf y}),...,\alpha_K({\bf y})$, to be determined later,
\begin{align}
\label{eq:proposal}
\widetilde q_X\bigl(\bx;\bs{\alpha}({\bf y})\bigr) &=\sum_{k=1}^{K} \alpha_{k}({\bf y}) \; g_k(\mathbf{x})= p_X(\mathbf{x}) \sum_{k=1}^{K} \frac{\alpha_{k}}{m_k}  \, 1_{R_k}(\bx),
\end{align}
where $\bs{\alpha}({\bf y})=\bigl(\alpha_{1}({\bf y}),\dots,\alpha_{K}({\bf y})\bigr)$ are the mixture coefficients of the proposal distribution (which are non-negative and add to one, \textit{i.e.},  $\bs{\alpha}({\bf y})$ belong to the $(K-1)$-dimensional probability simplex $\Delta^{(K-1)}$), and $1_{A}$ denotes the indicator function of some set $A$, that is, $1_{A}({\bf x}) = 1$, if ${\bf x}\in A$, and 0 otherwise. The notation $\bs{\alpha}({\bf y})$ is used to stress that these weights will be adapted as a function of ${\bf y}$. The resulting SNIS weights for some sample ${\bf x}$ derived from $\widetilde q_X$ are given by
\begin{equation}
\label{eq:weights}
w(\mathbf{x}) = \frac{l_{\bf y}(\bx)\, p_X(\mathbf{x})}{\widetilde q_X(\bx;\bs{\alpha}({\bf y}))} =\frac{l_{\bf y}(\bx)}
     {\sum_{k=1}^{K} \frac{\alpha_{k}({\bf y})}{m_k}  1_{R_k}(\mathbf{x} )}.
\end{equation}
The choice of optimal weights $\bs{\alpha}({\bf y})$ will be discussed later in this section, as well as how to sample from $ \widetilde q_X(\bx;\bs{\alpha}({\bf y}))$ without knowing $p_X$.

This proposed SNIS estimator has a few distinctive features, namely: \textbf{(i)} knowledge of the marginals  $p_X(\bx)$ is $p_Y(\by_i)$ is not needed; \textbf{(ii)} any likelihood function $l_{\bf y}(\bx)$ can be used; \textbf{(iii)} the samples $ \mathbf{x}_j \sim \widetilde q_X(\cdot;\bs{\alpha}({\bf y}))$ can be easily obtained from the external dataset {as} seen below.
	
In earlier work \cite{niknejad_classg_2017,niknejad_classp_2017}, we proposed IS-based methods for denoising.  However, those methods are  quite different from the one herein proposed: in those methods,  a distribution for each patch is selected from a set of learned distributions, and then external NLM is used.


\subsection{Optimizing the Proposal Distribution}
We now address the  setting of $\bs{\alpha}({\bf y})$ in the  proposal distribution \eqref{eq:proposal}. Our approach is based on   \eqref{eq:optsnis}, which provides the optimal proposal distribution, and on a similarity measure between two probability distributions. Since the target distribution is the posterior distribution $p_{X|Y}({\bf x}\,|\mathbf{y})$, the optimal sampling distribution is
\begin{equation}
\label{eq:opt_q_xc}
q_{X}^{*}(\mathbf{x}) \propto |\mathbf{x}_c-\widehat{\mathbf{x}}_{c}| \; p_{X|Y}({\bf x}\,|\mathbf{y}).
\end{equation}
The optimal proposal distribution $q_X^{*}$ depends on $\widehat{\mathbf{x}}_{c}$, which we do not have, as it is precisely the object of the estimation procedure.
Later in this section, we will discuss an alternating minimization approach to address this issue. For now, assume that an estimate $\widehat{\mathbf{x}}_{c}$ is available. 

A natural criterion to adjust the weight vector $\bs{\alpha}({\bf y})$ is to minimize some \textit{distance} measure between  $\widetilde q_X(\cdot\,;\bs{\alpha})$ and $q^*$. An obvious choice would be the Kullback-Leibler divergence \cite{kullback_information_1951}; however, it is not symmetric and its computation is not straightforward for some distributions. Here, we  use the squared Hellinger distance \cite{le_asymptotics_2012}, which, given  two probability density functions  $q$ and $p$,  is defined as
\begin{equation}
\label{eq:hellinger}
H^2(p,q)= 1-{\int \sqrt{p(\mathbf{x})  q(\mathbf{x})} \ d\mathbf{x}}.
\end{equation}
The Hellinger distance is a metric satisfying all the corresponding properties (namely, symmetry and triangle inequality). The application of the Hellinger distance to our setup yields 
\begin{eqnarray}
\label{eq:disprobopt}
\widehat{\boldsymbol{\alpha}} = \argmin_{\boldsymbol{\alpha}\in\Delta^{K-1}} {H^2(q_X^{*}, \widetilde q_X(\cdot \, ;\boldsymbol{\alpha}))} \nonumber \\ =\argmax_{\boldsymbol{\alpha}\in\Delta^{K-1}} \int \sqrt{\widetilde q_X(\mathbf{x};\boldsymbol{\alpha}) q^*(\mathbf{x})} \ d\mathbf{x}.
\end{eqnarray}
It is worth mentioning that Bhattacharyya distance (another distance measure for probability density functions) leads to the same  optimization problem \eqref{eq:disprobopt}, as it is defined as $B(p,q)=-\ln {\int \sqrt{p(\mathbf{x})  q(\mathbf{x})} \ d\mathbf{x}}$ \cite{bhattacharyya_measure_1943}.

By inserting \eqref{eq:proposal} and \eqref{eq:opt_q_xc} into \eqref{eq:disprobopt}, we obtain
\begin{equation}
\widehat{\boldsymbol{\alpha}} = \argmax_{\boldsymbol{\alpha} \in \Delta^{K-1}}       
       \!\! \int_{\mathbb{R}^n}  \!\!
       \biggl(\big|\mathbf{x}_c-\widehat{\mathbf{x}}_{c}\big|\,
     	   {l_{\bf y}(\bx)\,} \!\!
     	   {}\sum_{k=1}^{K} \!\! \frac{\alpha_k}{m_k}  
     1_{\mathcal{X}_k}(\mathbf{x}) \biggr)^{\tfrac{1}{2}} \!\! p_{X}(\mathbf{x}) d\mathbf{x}.\label{eq:integral1}
\end{equation}

To approximate the integral in \eqref{eq:integral1}, we resort to MC sampling from $p_X$, by using M samples, ${\bf x}_1,...,{\bf x}_M,$ from the available external dataset, yielding
\begin{equation}
\widehat{\boldsymbol{\alpha}}=\argmax_{\boldsymbol{\alpha} \in \Delta^{K-1}} \sum_{s=1}^{M}  \biggl(\big|(\mathbf{x}_{s})_c-\widehat{\mathbf{x}}_{c}\big|\,{l_{\bf y}(\bx_s)} \sum_{k=1}^{K} \frac{\alpha_k}{m_k}  1_{\mathcal{X}_k}(\mathbf{x}_s)\biggr)^{1/2}.
\label{eq:obtalpha}
\end{equation}
Notice that the samples ${\bf x}_1,...,{\bf x}_M$ can be partitioned  according to which cluster $\mathcal{X}_1,...,\mathcal{X}_K$ each one belongs to. Furthermore, noticing that if ${\bf x}_s \in \mathcal{X}_j$, then $1_{\mathcal{X}_j}(\mathbf{x}_s) = 1$, whereas $1_{\mathcal{X}_r}(\mathbf{x}_s) = 0$, for $r\neq j$. This allows re-writing \eqref{eq:obtalpha} as
\begin{equation}
\label{eq:optalpha2}
\widehat{\boldsymbol{\alpha}}=\argmax_{\boldsymbol{\alpha} \in \Delta^{(K-1)}} \sum_{k=1}^{K} \; \Bigl(\frac{\alpha_k}{m_k}\Bigr)^{\tfrac{1}{2}}\!\!
\sum_{s: {\bf x}_s \in \mathcal{X}_k} \biggl(\big| (\mathbf{x}_{s})_c -\widehat{\mathbf{x}}_{c}\big|\,{l_{{\bf y}}(\mathbf{x}_{s})} \biggr)^{\tfrac{1}{2}}.
\end{equation}

Before proceeding, recall that $m_k$ is given by \eqref{eq:m_k}, thus its MC-based estimate is simply $\widehat{m}_k = |\mathcal{X}_k|/L$. Plugging this estimate in the previous expression yields
\begin{equation}
\label{eq:optalpha3}
\widehat{\boldsymbol{\alpha}}=\argmax_{\boldsymbol{\alpha} \in \Delta^{(K-1)}} \sum_{k=1}^{K} \; \sqrt{\alpha_k} \; b_k,
\end{equation}
where 
\begin{equation}
b_k = \biggl( \frac{1}{ |\mathcal{X}_k|}\biggr)^{1/2}\!\!
\sum_{s: {\bf x}_s \in \mathcal{X}_k} \biggl(\big| (\mathbf{x}_{s})_c -\widehat{\mathbf{x}}_{c}\big|\,{l_{{\bf y}}(\mathbf{x}_{s})} \biggr)^{1/2}.
\end{equation}

Finally, the optimal solution of \eqref{eq:optalpha3} is
\begin{equation}
\label{eq:optcoe}
\widehat{\alpha}_k=\frac{b_k^2}{ \sum_{k=1}^K b_k^2}, \qquad  k=1,\ldots,K.
\end{equation}
To see why this is so, consider the following change of variables: $\beta_k = \sqrt{\alpha_k}$, thus $\alpha_k = \beta_k^2$. With this change of variable, problem \eqref{eq:optalpha3} becomes 
\begin{equation}
\widehat{\boldsymbol{\beta}}=\argmax_{\boldsymbol{\beta} \in S^{(K-1)}} \sum_{k=1}^{K} \; \beta_k \; b_k = \argmax_{\boldsymbol{\beta} \in S^{(K-1)}} \boldsymbol{\beta}^T{\bf b},
\end{equation}
where $S^{K-1}$ denotes  the unit-radius sphere in $\mathbb{R}^n$. The solution of this problem is well known to be the normalization (to unit norm) of ${\bf b}$, that is,
$\widehat{\boldsymbol{\beta}} = {\bf b}/\|{\bf b}\|_2$. Finally, inverting the change of variable yields the solution \eqref{eq:optcoe}.

Finally, because  the optimal $\widehat{\boldsymbol{\alpha}}$ does depend on $\bf y$ (because $\bf b$ depends on $\bf y$), we will recover the notation used in the previous subsection and refer to it as $\widehat{\boldsymbol{\alpha}}({\bf y})$.

\subsection{Sampling and Weighting}
After the optimal $\widehat{\boldsymbol{\alpha}}({\bf y})$ has been obtained as described in the previous subsection, we now explain how to obtain $N$ samples from the optimized proposal distribution
\begin{equation}
\label{eq:proposal2}
\widetilde q_X\bigl(\bx;\widehat{\bs{\alpha}}({\bf y})\bigr) =\sum_{k=1}^{K} \widehat{\alpha}_{k}({\bf y}) \; g_k(\mathbf{x}).
\end{equation}
Recall that a sample from this finite mixture can be obtained by first sampling from a categorical variable with probabilities $\widehat{\alpha}_{1}({\bf y}),...,\widehat{\alpha}_{K}({\bf y})$, and then sampling from the selected component. If the number of samples $N$ is large, this is approximately equivalent to obtaining  $N_k = \mbox{round}(\widehat{\alpha}_{k}({\bf y})\; N)$ samples from each component $g_k$, for $k=1,...,K$. The samples from the $k$-component are simply obtained by randomly sampling from the cluster of clean patches $\mathcal{X}_k$.

Computing the weight of each sample according to \eqref{eq:weights} is very simple, since only one of the terms in the sum in the denominator is non-zero: the one corresponding the cluster from which that sample was obtained. That is, 
\begin{equation}
{\bf x} \in \mathcal{X}_k \Rightarrow w({\bf x}) = \frac{m_k \; l_{\bf y}({\bf x})}{\widehat{\alpha}_{k}({\bf y})}.
\label{eq:snisweight}
\end{equation}

\subsection{Dealing with the Unknown $\widehat{\mathbf{x}}_{c}$}
As mentioned before, and is obvious in \eqref{eq:integral1}--\eqref{eq:optalpha2}, obtaining the optimal  $\widehat{\boldsymbol{\alpha}}({\bf y})$ requires  knowing $\widehat{\mathbf{x}}$, which is precisely the target of the estimation problem. To tackle this issue, we use an iterative approach that alternates between the following two steps, after initializing the estimate $\widehat{\mathbf{x}}_{c}$ with the noisy observation: \textbf{(i)} $\widehat{\boldsymbol{\alpha}}({\bf y})$ is computed via \eqref{eq:optcoe}; \textbf{(ii)} the estimate  $\widehat{\mathbf{x}}_{c}$ is updated via SNIS. These two steps repeated until some convergence criterion is satisfied.

\subsection{Full Patch Estimation}
We now consider a variant of the method above proposed that restores the whole patch, instead of just the central pixel thereof. This variant is faster and yields improved performance for the same running time. Results with the two implementations are provided in Section \ref{sec:experiments}.

Inspired by the patch-based approaches \cite{2007_dabov_bm3d,2006_elad_denoisesparse}, the noisy image is first divided into overlapping patches\footnote{In addition, we will test different strides (i.e., shifts between consecutive extracted patches), which control the trade-off between time complexity and denoising performance.}, then, instead of restoring merely the central pixel, the whole patch is denoised; finally, the patches are returned to the original positions and are averaged in overlapping pixels.  In the restoration step, we extend the central pixel estimate  \eqref{eq:isavg}  to the whole patch.Notice that the estimate in \eqref{eq:isavg} is still valid if $(\mathbf{x}_{j})_c$ is  replaced by $(\mathbf{x}_{j})_d$, where $d$ denotes any pixel index in a patch, not necessarily the central one. This modification raises an issue: the optimal sampling distribution in \eqref{eq:optsnis} is only valid if the range of function $f$ is $\mathbb{R}$, thus it is not directly applicable if the central pixel is replaced by the whole patch. We address this issue by choosing $\boldsymbol{\alpha}$ to maximizes the similarity measures for all pixels in a patch, on average, \textit{i.e.},
\begin{equation}
\widehat{\boldsymbol{\alpha}}=  \argmax_{\boldsymbol{\alpha} \in \Delta^{K-1}}       
       \sum_{d=1}^{n} \!\! \int_{\mathbb{R}^n}  \!\!
       \biggl( \!\! \big|(\mathbf{x})_d-\widehat{\mathbf{x}}_{d}\big|\,
     	   {l_{\bf y}(\bx)\,} \!\!
     	   {}\sum_{k=1}^{K} \!\! \frac{\alpha_k}{m_k}  
     1_{\mathcal{X}_k}(\mathbf{x}) \!\! \biggr)^{\tfrac{1}{2}} \!\!\! p_{X}(\mathbf{x}) d\mathbf{x}. \nonumber
\end{equation}
Following the same rational discussed in the previous section, and a simple rearrangement of summations, in this case $\boldsymbol{\alpha}$ is given by \eqref{eq:optcoe}, where
$$
b_k=\biggl( \frac{1}{ |\mathcal{X}_k|}\biggr)^{1/2}  \sum_{d=1}^{n} \,
\sum_{s: {\bf x}_s \in \mathcal{X}_k} \biggl(\big| (\mathbf{x}_{s})_d -\widehat{\mathbf{x}}_{d}\big|\,{l_{{\bf y}}(\mathbf{x}_{s})} \biggr)^{1/2}.
$$
The motivation for a whole-patch approach is that the computational bottleneck of the proposed method (as in NLM), is the computation of the likelihood function. Using the whole-patch procedure, computing the weights for the whole patch requires the same number of computations of the likelihood as for just the central pixel. Furthermore, by adjusting the stride (displacement between consecutive patches), it is possible to control the trade-off between computational cost and performance. In Section \ref{sec:experiments}, we report  result with both the whole-patch and central-pixel method. The algorithm with the central-pixel-based method is shown in Fig. \ref{fig:cppropsed}, while Figure \ref{fig:propsed} summarizes the proposed whole-patch scheme.

\subsection{Practical implementation}
For clustering the external dataset, any (hard clustering) algorithm, such as $k$-means, can be used, since the proposed restoration algorithm does not depend critically on the clustering. In our implementation, we use the \textit{classification-EM} (CEM) algorithm \cite{Celeux}, which fits $K$ multivariate Gaussian distributions to the data and considers the samples assigned to each multivariate distribution as a cluster. 

Based on the above considerations, it can be seen that there is no need for a parametric form of the mixture distribution  $\widetilde{q} (\cdot\,;\bs{\alpha})$ or even the distribution of natural images $p_X$. The proposed  approach may be seen as new general method based on IS, which, unlike other methods such as \cite{2015_koblents_population, cornuet_adaptive_2012}, does not require any parametric proposal distribution, and can potentially be used with other sources of data, if available. Another important feature of our proposed method is that it requires merely the evaluation of the likelihood, which is usually (assumed) known and easy to computed \cite{foi_practical_2008}. 

Implementation of the proposed method is computationally expensive if all the samples in the external dataset are used for obtaining the coefficients $\boldsymbol{\alpha}$, and estimation of the clean patch. However, as shown in the result section, a very limited number of samples $M$ (less than $1 \%$ of the whole dataset) suffices to obtain a good estimate of $\boldsymbol{\alpha}$. Our approach then uses a limited  number $N$ of so-called \textit{important} samples, derived from the proposal distribution to estimate $\hat{\mathbf{x}}_i$. So, the total number of samples (\textit{i.e.} $N+M$) is much smaller than the what is typically used in external NLM. We defer more discussion on the computational complexity to Section \ref{sec:experiments}. We will also show that the proposed SNIS approach performs better than similar efficient approaches for the large-scale Gaussian denoising, such as MC-NLM \cite{2014_chan_monte1}, which uses the sampled patches based on the sampling distribution obtained by the large deviation theories.

\begin{figure}
\centering
\begin{tabular}{|p{8.2cm}|}
\hline
\begin{itemize}[leftmargin=*]
\item For each pixel $i$ in the image
\begin{itemize}[leftmargin=*]
\item Extract the patch $\mathbf{y}_i$ with the pixel in center, and iterate for $L$ times:
\begin{itemize}[leftmargin=*]
\item Compute the mixture coefficients $\boldsymbol{\alpha}(\mathbf{y}_i)$ by \eqref{eq:optcoe}.
\item Extract the patches $\mathbf{x}_j$'s from the mixture distribution $\widetilde q_X\bigl(\bx;\bs{\alpha}({\bf y}_i)\bigr)$.
\item Estimate the central pixel by the weighted average \eqref{eq:isavg}.
\end{itemize}
\end{itemize}
\end{itemize}
 \\
\hline
\end{tabular}
\caption{The algorithm of the proposed SNIS method for central-pixel estimation.}
\label{fig:cppropsed}
\end{figure}

\begin{figure}
\centering
\begin{tabular}{|p{8.2cm}|}
\hline
\begin{itemize}[leftmargin=*]
\item Divide noisy image into overlapping patches $\mathbf{y}_i,\; i=1,\ldots I$.
\item For each patch $\mathbf{y}_i$, iterate for $L$ times:
\begin{itemize}[leftmargin=*]
\item Compute the mixture coefficients $\boldsymbol{\alpha}$ by \eqref{eq:optcoe}.
\item Extract the patches $\mathbf{x}_j$'s from the mixture distribution $\widetilde q_X\bigl(\bx;\bs{\alpha}({\bf y}_i)\bigr)$.
\item Compute the weighted average of patches using \eqref{eq:isavg} for all pixels of the patch.
\end{itemize}
\item Return the restored patches to the original position in the image and average in the overlapping pixels.
\end{itemize}
 \\
\hline
\end{tabular}
\caption{The algorithm of the proposed SNIS method for whole-patch estimation.}
\label{fig:propsed}
\end{figure}

\section{Experimental results}
\label{sec:experiments}
In this section, we assess the performance of the proposed method on both class-specific and large-scale generic image datasets.

\subsection{Class-specific image restoration}
In this subsection, we apply the proposed  method to several problems where a dataset of the same class of the noisy image is available; in fact, in many applications, the image class is known or can be obtained (\textit{e.g.}, an image of a face, a fingerprint, text, etc.). This scenario has recently received considerable attention \cite{remez_deep_2017, 2015_luo_adaptive,anwar_category_2017,2016_teodoro_image}. 

The first set of experiments addresses denoising, where the observed image is contaminated by either Gaussian or Poissonian noise. The second set of experiments addresses image inpainting, \textit{i.e.}, some pixels in the image are missing. The combination of noise with missing pixels is also considered. The Gore face dataset \cite{2012_peng_rasl} and the MNIST handwritten digits dataset \cite{lecun_gradient_1998} are adopted as external datasets. For the text dataset, we extracted images from different parts of typed text documents. For the face and text dataset, $5$ images are considered as the test image and the others are used as external. For the MNIST handwritten dataset, we used the splitting into training and testing that is provided in the corresponding website. Regarding the proposed method, we have found that, to achieve  the best performance, different number of clusters are needed for different datasets. For example, for face, text, and handwritten digit datasets, we use $20$, $30$, and $50$ clusters of $9\times9$ patches were used, respectively. 

All the results reported for the class-specific cases were obtained using the whole-patch-based algorithm in Figure \ref{fig:propsed}, with the following parameter settings.  A small number of random samples for each of the two stages (\textit{i.e.}, $N$ and $M$) is used. Specifically, we set $M=900$, which is less than $1\%$ of the samples in each external datasets and $N = 300$. Therefore, for each degraded patch, a total of $1200$ sample patches are used, which involves a computational complexity roughly similar to an internal non-local denoising procedure, with the patches constrained approximately to a $35\times35$ search window. The number of iterations $L$ in the alternating minimization approach was set to 3. The stride for selecting patches in the noisy image is set to 2, to further reduce the computational cost. We empirically found that this value is sufficient to achieve competitive denoising results, while keeping the computational cost reasonably low. More details about the computational cost of the proposed algorithm will be discussed in the subsection related to the large-scale generic dataset.

\subsubsection{Gaussian denoising}
\begin{figure}[!t]
\centering
\subfloat[]{\includegraphics{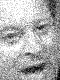}%
\label{a)}}~
\subfloat[]{\includegraphics{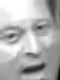}%
\label{(b)}}~
\subfloat[]{\includegraphics{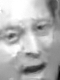}%
\label{(c)}}~
\\
\subfloat[]{\includegraphics{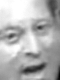}%
\label{(d)}}~
\subfloat[]{\includegraphics{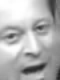}%
\label{(e)}}~
\subfloat[]{\includegraphics{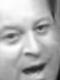}%
\label{(f)}}
\hfill
\caption{Example of denoising for a face image in the Gore dataset ($\sigma=30$): (a) noisy image; (b) BM3D (PSNR=29.80\,dB); (c) EPLL (PSNR=28.28\,dB); (d)~Class specific EPLL (PSNR=29.98\,dB); (e) Class-specific denoising in \cite{2015_luo_adaptive}~(PSNR=32.98\,dB); (f) this work (PSNR=34.11\,dB).}
\label{fig:faceg}
\end{figure}

The first set of experiments addresses image denoising under i.i.d.  Gaussian additive noise, \textit{i.e.}, the observation model is given by  \eqref{eq:LIP} with ${\bf H} = {\bf I}$, thus the likelihood is \eqref{eq:glik}. Table \ref{tb:gaussian} shows denoising \textit{peak-signal-to-noise-ratio} (PSNR) results (in dB) for various values of the noise standard deviation $\sigma$ and for the MNIST and text datasets. In addition to the results for the proposed SNIS method, the table also shows results for the methods EPLL (generic)  and BM3D (generic) and  \cite{2015_luo_adaptive} (class-specific). The proposed method outperforms, not only methods designed to generic images, but also the method proposed in \cite{2015_luo_adaptive}, which is class-specific. An example of a denoising result for the face dataset is shown in Figure \ref{fig:faceg}.

\begin{table*}[]
\renewcommand*{\arraystretch}{1.3}
\centering
\caption{Gaussian denoising:  PSNR (dB) averaged over 5 test images. Datasets: MNIST \cite{2012_peng_rasl} and  text.}
\label{tb:gaussian}
\begin{tabular}{l|l|l|l|l|l|l|l|l|}
\cline{2-9}
 & \multicolumn{2}{c|}{$\sigma=20$} & \multicolumn{2}{c|}{$\sigma=30$} & \multicolumn{2}{c|}{$\sigma=40$} & \multicolumn{2}{c|}{$\sigma=50$} \\ 
 \cline{2-9} 
 & \multicolumn{1}{|c|}{MNIST} & \multicolumn{1}{|c|}{text}  & \multicolumn{1}{|c|}{MNIST} & \multicolumn{1}{|c|}{text}  & \multicolumn{1}{|c|}{MNIST} & \multicolumn{1}{|c|}{text}  & \multicolumn{1}{|c|}{MNIST} & \multicolumn{1}{|c|}{text} \\   \hline  
 \multicolumn{1}{|c|}{BM3D}& 28.30& 28.13&25.39&24.95&21.81&22.55&26.98&20.91  \\ \hline
 \multicolumn{1}{|c|}{EPLL}&28.51&28.15&25.64&25.21&	22.07&23.15&22.77&21.72 \\ \hline
\multicolumn{1}{|c|}{Luo et. al. \cite{2015_luo_adaptive}} &27.03&27.52&26.34&27.44&24.81&26.29&24.79&25.02 \\ \hline
\multicolumn{1}{|c|}{SNIS}& \bf{29.46}&\bf{29.98}&\bf{28.62}&\bf{29.19}&\bf{26.01}&\bf{28.31}&\bf{27.75}&\bf{27.29} \\ \hline
\end{tabular}
\end{table*}

\subsubsection{Poisson Denoising}
In these experiments, we generate images corrupted by Poissonian noise, according to model \eqref{eq:poisson} with ${\bf H} = {\bf I}$.  Table \ref{tb:poissonre} shows denoising PSNR results averaged over 5 test images for the above-mentioned face and text datasets. The performance of SNIS is compared with those of NL-PCA \cite{2014_salmon_poisson}, VST+BM3D \cite{2011_makitalo_optimal}, and P4IP \cite{2016_rond_poisson}, in low SNR regimes. All those methods are generic, as we know of no other class-specific Poisson denoising method performing well on our dataset\footnote{There is one method based on deep networks \cite{remez_deepp_2017}, which is learned on much larger datasets, such as ImageNet.}. The results show that SNIS noticeably outperforms other methods. An example of denoising of 5 digits from the MNIST dataset is shown in Figure \ref{fig:poissonde}.

\begin{table*}[]
\centering
\renewcommand*{\arraystretch}{1.3}
\caption{Class-specific Poisson denoising for different peak values shown in the first row.  The reported PSNR (dB) is averaged over 5 test images.}
\label{tb:poissonre}
\begin{tabular}{l|l|l|l|l|l|l|l|l|}
\cline{2-9}
 & \multicolumn{2}{c|}{$10$} & \multicolumn{2}{c|}{$5$} & \multicolumn{2}{c|}{$2$} & \multicolumn{2}{c|}{$1$} \\ 
 \cline{2-9} 
 & \multicolumn{1}{|c|}{Face} & \multicolumn{1}{|c|}{text}  & \multicolumn{1}{|c|}{Face} & \multicolumn{1}{|c|}{text}  & \multicolumn{1}{|c|}{Face} & \multicolumn{1}{|c|}{text}  & \multicolumn{1}{|c|}{Face} & \multicolumn{1}{|c|}{text} \\   \hline  
 \multicolumn{1}{|c|}{NL-PCA}& 25.01&22.16& 23.80&19.66&22.87&14.92&19.69 &12.70  \\ \hline
 \multicolumn{1}{|c|}{VST+BM3D}&25.41&23.15&24.79&20.96&23.70&16.29&20.80&13.89 \\ \hline
\multicolumn{1}{|c|}{P4IP} &25.84&23.51&24.88&21.19&23.78&17.22&20.03&14.12 \\ \hline
\multicolumn{1}{|c|}{SNIS}& \bf{27.40}&\bf{24.32}&\bf{25.78}&\bf{23.83}&\bf{23.95}&\bf{22.90}&\bf{21.31}&\bf{21.02 }\\ \hline
\end{tabular}
\end{table*}

\begin{figure}
\centering
\renewcommand*{\arraystretch}{1.3}
\caption{Examples of denoising of digits from the MNIST dataset contaminated with Poisson noise. The peak value of the original image is 2.}
\label{fig:poissonde}
\begin{tabular}{>{\centering\arraybackslash}m{1cm} >{\centering\arraybackslash}m{.8cm} >{\centering\arraybackslash}m{.8cm} >{\centering\arraybackslash}m{.8cm} >{\centering\arraybackslash}m{.8cm} >{\centering\arraybackslash}m{.8cm}}
Original image&\includegraphics{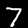}&\includegraphics{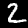}&\includegraphics{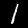}&\includegraphics{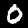}&\includegraphics{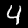} \\
NL-PCA&\includegraphics{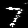}&\includegraphics{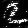}&\includegraphics{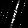}&\includegraphics{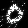}&\includegraphics{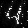} \\
P4IP&\includegraphics{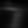}&\includegraphics{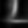}&\includegraphics{images/poisson_mnist/nlpca-1447}&\includegraphics{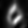}&\includegraphics{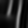} \\
SNIS&\includegraphics{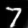}&\includegraphics{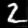}&\includegraphics{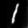}&\includegraphics{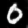}&\includegraphics{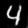} \\
\end{tabular}
\end{figure}

\subsubsection{Image Inpainting in Additive Gaussian Noise}
\begin{figure*}
\subfloat[]{\includegraphics[width=.33\linewidth]{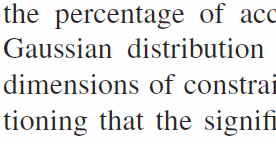}%
\label{a)}}~
\subfloat[]{\includegraphics[width=.33\linewidth]{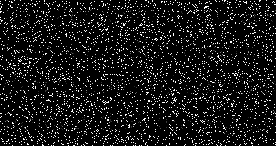}%
\label{b)}}~
\subfloat[]{\includegraphics[width=.33\linewidth]{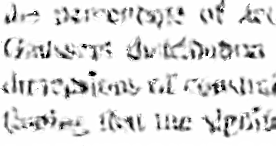}%
\label{(c)}}~
\\
\subfloat[]{\includegraphics[width=.33\linewidth]{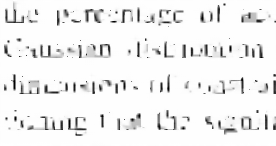}%
\label{(d)}}~
\subfloat[]{\includegraphics[width=.33\linewidth]{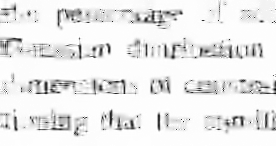}%
\label{(e)}}~
\subfloat[]{\includegraphics[width=.33\linewidth]{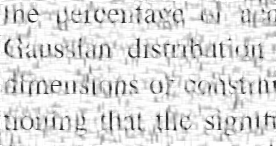}%
\label{(f)}}~
\caption{Image inpainting results: (a) original image; (b)  degraded image (10\% of pixels are available); (c)  kernel regression (PSNR=13.91\,dB); (d) FOE (PSNR=15.24\,dB); (e) EPLL (PSNR=14.85\,dB); (f) SNIS (PSNR=16.63\,dB).}
\label{fig:inttext}
\end{figure*}

Image inpainting is the problem of recovering images in which some pixels are missing. In this case, $\mathbf{H}$ is a diagonal matrix in which the diagonal entries are either zero or one, corresponding to the missing or available pixels, respectively. In the noiseless case, the value of $\sigma$ in \eqref{eq:glik} can be  taken very small. However, for a very small value of $\sigma$, the likelihood function can become numerically zero for some observed patches $\mathbf{y}_i$; it may even happen that for some noisy patches, all the sampled patches have zero likelihood. In this case, the estimate in \eqref{eq:glik} is not defined. In order to avoid this numerical problem, for each observed patch, the value of $\sigma$ is initialized to a very small value, and for the patches in which the obtained weights are all zero, $\sigma$ iteratively increased to achieve at least one non-zero weight. The results of image inpainting for different percentages of randomly missing pixels are reported in Table \ref{tb:inttb}. The proposed method is compared to EPLL \cite{2011_zoran_learning}, kernel regression (KR) \cite{takeda_kernel_2007}, and the field of experts (FoE) \cite{2005_roth_fields} methods, all of which are generic. An example of a restored text image obtained by the proposed method is compared to other methods in Figure \ref{fig:inttext}.

\begin{table}[]
\centering
\renewcommand*{\arraystretch}{1.3}
\caption{Image inpainting results for the Gore face  and MNIST datasets,  different percentage of randomly available pixels. The reported PSNR is averaged over five test images. }
\begin{tabular}{l|l|l|l|l|l|l|}
\cline{2-7}
 & \multicolumn{2}{c|}{$50\%$} & \multicolumn{2}{c|}{$20\%$} & \multicolumn{2}{c|}{$10\%$} \\ 
 \cline{2-7} 
 & \multicolumn{1}{|c|}{face} & \multicolumn{1}{|c|}{MNIST}  & \multicolumn{1}{|c|}{face} & \multicolumn{1}{|c|}{MNIST}  & \multicolumn{1}{|c|}{face} & \multicolumn{1}{|c|}{MNIST} \\   \hline  
 \multicolumn{1}{|c|}{KR}& 32.79 & 18.26 & 30.78 & 16.58& 24.51 &13.82\\ \hline
 \multicolumn{1}{|c|}{FOE}&41.72&  26.02 & 31.21& 18.58 &26.18 &15.60\\ \hline
\multicolumn{1}{|c|}{EPLL} &39.36&24.50&30.25 &17.98 & 25.20&15.15 \\ \hline
\multicolumn{1}{|c|}{SNIS}& \bf{48.30}&\bf{26.33}&\bf{37.47}&\bf{22.22}&\bf{29.26}&\bf{17.20}\\ \hline
\end{tabular}
\label{tb:inttb}
\end{table}

\begin{figure}
\centering
 \subfloat[]{\includegraphics[]{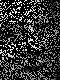}%
\label{a)}}~
\subfloat[]{\includegraphics[]{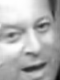}%
\label{b)}}\\
 \subfloat[]{\includegraphics[width=.47\linewidth]{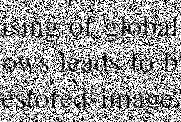}%
\label{a)}}~
\subfloat[]{\includegraphics[width=.47\linewidth]{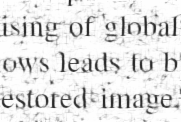}%
\label{b)}}~
\caption{Examples of recovering images with missing pixels and contaminated by noise using SNIS. (a) Degraded image: $70\%$ randomly available pixels with additive i.i.d. Gaussian noise with standard deviation 15. (b) Restored image (PSNR=33.26\,dB). (c) Degraded image with peak of 5 contaminated by Poisson noise and observation of $80\%$ of pixels observed (PSNR=4.92 \,dB). (d) Restored image (PSNR=20.18\,dB)}.
\label{fig:intgaussian}
 \end{figure}

Here, we evaluate our proposed method in a more general observation models which is the combination of noise and missing pixels. An example of the restored images is shown in Fig.~\ref{fig:intgaussian} for the Gaussian and Poisson noise models. For the Poisson case, some weights are ambiguous as the term $(({\mathbf{H}\mathbf{x})_{l}})^{(\mathbf{y})_{l}}$ in \eqref{eq:poisson} becomes zero in both base and exponent. In this case, we set these entries to $1$ as no information is available for the missing pixel. Any other constant value is also possible, since it cancels out in the weighted average. 

 \subsection{Generic image denoising}
If the class of image is known, the prior $p_X$ is well-adapted to the noisy image, resulting in state-of-the-art performance. However, for generic images, the prior from natural image patches is not specifically adapted to the image. As discussed in \cite{2011_levin_natural}, the performance of BM3D denoising is close to the MMSE estimate for generic images. However, here we focus on another major challenge of using sampling methods in the generic datasets which is the large number of samples in the external datasets. Unlike in the class-specific case, where the number of samples in the external datasets is limited, generic images require a very large dataset to achieve good denoising performance. In this case, sampling methods such as external NLM become computationally very expensive \cite{2011_levin_natural}.

It is well-known that the computational cost of the NLM algorithm for denoising one pixel is $O(Sn)$  where $S$ is the number of patch samples and $n$ is the number of pixels in a patch \cite{2014_chan_monte}. So, the computational complexity can roughly be measured by the number of processed patches from the external dataset. As a result, a set of proposed methods limits the number of processed samples from the dataset to speed up denoising. Our proposed method belongs to this group and, in this section, we compare our method with other methods in this class. One approach would be simply selecting randomly a specific number of samples $S$  from the dataset and then compute the weighted average \eqref{eq:isavg} to estimate the pixel. We call this method \textit{uniform sampling} (following \cite{2014_chan_monte}). Another approach proposed is MC-NLM \cite{2014_chan_monte}, where a Bernoulli sampling probability is assigned to each patch based on large-deviation theory techniques. The patches are then selected based on this sampling distribution and the weighted average is computed. In our method, the number of processed patches is computed  as the sum of the number of patches processed for the mixture coefficients $M$ and the number of patches used for estimation, $N$, \textit{i.e.}, $S=N+M$. For the MC-NLM, apart from the number of samples, the complexity of obtaining the sampling distribution for each patch should also be considered.

\begin{figure}
	\centering
\subfloat[]{\includegraphics[width=.48\linewidth]{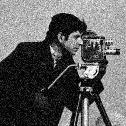}%
\label{a)}}~
\subfloat[]{\includegraphics[width=.48\linewidth]{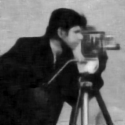}%
\label{b)}}\\
\subfloat[]{\includegraphics[width=.48\linewidth]{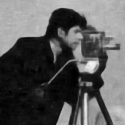}%
\label{c)}}~
\subfloat[]{\includegraphics[width=.48\linewidth]{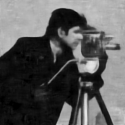}%
\label{d)}}~
\caption{Denoising example (Gaussian noise): (a) noisy image ($\sigma=30$); (b) uniform sampling: PSNR=23.08\,dB, CPU time = 93 seconds; (c) SNIS: PSNR=24.16\,dB, CPU time=94 seconds; (d) exact MMSE: PSNR=24.78\,dB, CPU time=3 hours.}
\label{fig:genericcman}
\end{figure}
\begin{figure}
	\centering
\subfloat[]{\includegraphics[width=.48\linewidth]{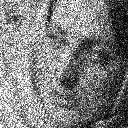}%
\label{a)}}~
\subfloat[]{\includegraphics[width=.48\linewidth]{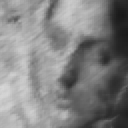}%
\label{b)}}\\
\subfloat[]{\includegraphics[width=.48\linewidth]{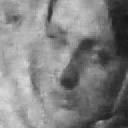}%
\label{c)}}~
\subfloat[]{\includegraphics[width=.48\linewidth]{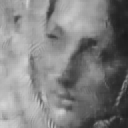}%
\label{d)}}~
\caption{Denoising a part of the Barbara image with Poisson noise: (a) noisy image (peak=$10$); (b) uniform sampling, PSNR=24.63\,dB,  CPU time = 157 seconds; (c) SNIS, PSNR=25.16\,dB, CPU time=158 seconds; (d) exact MMSE, PSNR=25.35\,dB, CPU time = 3.5 hours.}
\label{fig:genericbarbara}
\end{figure}

The experiments in this section use $2\times10^6$ external clean patches extracted from the generic image dataset in
\cite{martin_database_2001}. Figures \ref{fig:genericcman} and \ref{fig:genericbarbara} show two denoising examples for Gaussian and Poisson noise, respectively. In both cases, our method is compared to the uniform sampling and the exact MMSE \cite{2011_levin_natural}, with the following parameters. For SNIS, the patches in the external dataset are divided into $220$ clusters; we use two iterations with $M=2200$ and $N=300$. The number of samples for uniform sampling was set to $5000$ (the same total number of samples as our method). The exact MMSE estimate is equivalent to external NLM in the Gaussian noise case. It can be seen that in both experiments, our proposed method outperforms the uniform sampling by an noticeable margin, while the CPU time is roughly the same. The exact MMSE denoising, however, outperforms both methods but with a huge increase in  computational cost. It can also be seen that the performance of the proposed method with a limited number of samples is not far from the optimal MMSE.

\begin{figure*}
	\centering
\includegraphics[width=.7\linewidth]{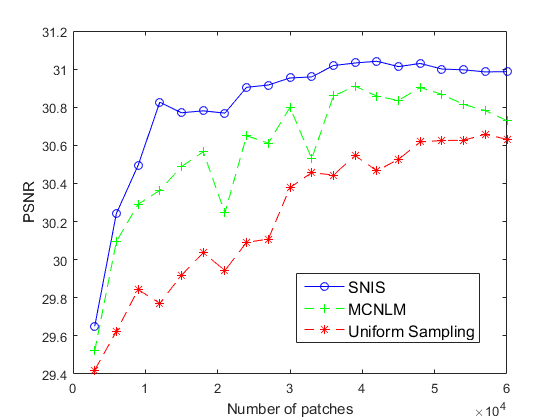}
\caption{Gaussian denoising of the central pixels of $2000$ noisy image patches ($\sigma = 18$) using a large-scale dataset. PSNR obtained by different methods as a function of the number of processed patches from the dataset.}
\label{fig:psnrplot}
\end{figure*}

In a final experiment, the discussed methods are used to estimate central pixels of $2000$ patches used in \cite{2014_chan_monte}, under Gaussian noise with $\sigma=18$. As discussed above, for the methods which use a subset of patch samples in the weighted average, the computational cost is determined by the number of samples used \footnote{For the MC-NLM method \cite{2014_chan_monte}, we neglect the computational complexity of obtaining the sampling distribution.}. Figure \ref{fig:psnrplot} plots PSNR as a function of the number of sampled patches, up to $6 \times 10^4$. This range of is suitable for computationally efficient denoising. Each point in the plot indicates one stage evaluation of the methods consisting of processing $3000$ patches. For the proposed method, the central pixel estimation algorithm in \ref{fig:cppropsed} is used, and the horizontal axis indicates the total number of patches processed, \textit{i.e.}, for obtaining the mixture plus the patches used for estimation. In each stage, $0.6$ of the total number of patches are used for updating the mixture coefficients $\boldsymbol{\alpha}$, and the others are used for updating the estimate. At each stage, the estimated pixel is replaced with $\hat{\mathbf{x}}_{c}$ in \eqref{eq:optalpha2} in order to obtain $\boldsymbol{\alpha}$. It can be seen that SNIS outperforms uniform sampling and MC-NLM for all the numbers of patches considered.

\section{Conclusion}
We introduced a new \textit{self-normalizing importance sampling} (SNIS) approach for image restoration, using samples from an external dataset of clean patches. We showed that the external non-local means (NLM) algorithm is a special case of the proposed method for Gaussian noise.  Our method has the following main advantages: a) it applies seamlessly to any data observation model for which the likelihood function can be computed, namely Gaussian and Poisson noise; b) it  is applicable to large-scale external datasets; and c) it yields  state-of-the-art results for tested class-specific datasets.
 
\bibliographystyle{IEEEtran}
\bibliography{ref}

\end{document}